\documentclass[letterpaper, 10 pt, conference]{ieeeconf}  
\IEEEoverridecommandlockouts                              
 
\usepackage{graphicx}   
\usepackage{subcaption} 
\usepackage{booktabs}   
\usepackage{multirow}   
\usepackage{amsmath} 
\usepackage{amssymb}  

\title{\LARGE \bf
Impedance Cloning: Learning Equilibrium Point Parameters \\ for Contact-Rich Manipulation
}

\author{Hayato Takahashi, Ryoga Oishi, Yuki Kasuga, and Toshiaki Tsuji%
\thanks{This work has been submitted to the IEEE for possible publication. Copyright may be transferred without notice, after which this version may no longer be accessible.}%
\thanks{The authors are with the Graduate School of Science and Engineering, Saitama University, Saitama, Japan
{\tt\small \{takahashi, oishi, kasuga\}@robotics.ees.saitama-u.ac.jp},
{\tt\small tsuji@ees.saitama-u.ac.jp}}%
}

\begin{document}

\maketitle

\begin{abstract}
 
Contact-rich manipulation requires robots to regulate force against surfaces whose geometry deviates unpredictably from training conditions. Trajectory-based imitation learning, which reproduces observable outputs, breaks down under such shifts. We propose Impedance Cloning, which instead imitates the biomechanical priors that generate motion---the stiffness and equilibrium point---and thereby passively absorbs contact uncertainty. Because these parameters encode intent rather than outcome, they generalize across surface geometries where trajectory reproduction does not. We extract them from bilateral teleoperation demonstrations via a particle filter without force/torque sensors and evaluate the framework on two CRANE-X7 manipulators. In a wiping task with joint-space actions, the trajectory-based baseline loses contact below $-6$\,cm, whereas the proposed method maintains a consistent 4--5\,N contact force above $-6$\,cm, with a gradual decrease below; with Cartesian-space actions, its force--height slope over 0 to $+8$\,cm is $0.13 \pm 0.03$\,N/cm, versus 0.34--0.83\,N/cm for fixed-impedance baselines. In a pick-and-place task with 10 diverse cups (100 trials), the proposed method succeeds in 84 trials, outperforming the fixed-impedance baseline (74/100) and performing comparably to a variable impedance control baseline (82/100) with one demonstration instead of ten. In a grasping task, the representation reduces torque tracking error with both ILBiT and Mamba backbones, confirming its generality across architectures.

\textit{Index Terms}---
Compliance and Impedance Control, Imitation Learning, Learning from Demonstration, Bioinspired Robot Learning.
\end{abstract}

\section{INTRODUCTION}
\label{sec:introduction}
Contact-rich manipulation requires a robot to regulate both position and force against unpredictable surface geometries. Current imitation learning methods---including ACT~\cite{zhao2023act},
Diffusion Policy~\cite{chi2023diffusion}, and teleoperation-based frameworks such as
Mobile ALOHA~\cite{fu2024mobilealoha}---define their action spaces over observable
states: end-effector position and force, or joint angles and torques. These representations reproduce
trained motions faithfully within the training distribution, but encode no intrinsic
compliance. For example, when the board height in a wiping task deviates from training conditions,
conventional trajectory-level policies either lose contact entirely or apply excessive force,
because the learned policy has no mechanism to convert positional error into adaptive contact force.
While data-driven variable impedance control methods exist, they often require estimating and re-identifying the environmental dynamics and stiffness for each new condition; therefore, generalization to unknown environmental changes remains difficult.
 
The Equilibrium Point~(EP) hypothesis~\cite{feldman1966functional, feldman1986, bizzi1992does} provides a
principled account of how the human motor system avoids this failure mode. Rather than
commanding torques directly, the central nervous system modulates two latent control
parameters---the virtual equilibrium point~$\theta^{\mathrm{eq}}$ and limb
stiffness~$K^{\mathrm{eq}}$---and relies on spring-like muscle mechanics to absorb
contact uncertainty passively. Position and force are observable \emph{outputs} of human motor planning, whereas stiffness and equilibrium point are its latent \emph{inputs}.
 
This distinction motivates our approach. Incorporating this biomechanical prior constructively, we redefine the action space of imitation learning in terms of EP parameters. Extracting both stiffness and equilibrium point simultaneously from only position and force observations is inherently an ill-posed problem. We make this estimation possible by combining real-world observation data collected via bilateral control with a particle filter that assumes continuous, small changes in the command parameters. The estimated $K^{\mathrm{eq}}$ and $\theta^{\mathrm{eq}}$ sequences are then either replayed directly as recorded (Motion Copy, in the replay-based experiments) or generated autonomously by a policy trained on them (ILBiT or Mamba, in the experiments with generative models). Because the robot imitates the generative parameters of human motor control rather than its observable outputs, it acquires the adaptive mechanism the human used to handle environmental uncertainty, not merely a record of what the human did under one specific condition. That is, imitation is carried out in the generative space rather than in the observable space.

We validate the framework across three tasks---wiping, pick-and-place, 
and grasping---using a bilateral CRANE-X7 setup with as few as 
10--16 demonstrations and no external force/torque sensors.
Notably, in the wiping task, whereas the trajectory-based baseline completely loses contact at a height variation of $-6$\,cm, the proposed method maintains a consistent contact force through passive compliance.
 
Our contributions are summarized as follows:
\begin{itemize}
    \item We propose \textit{Impedance Cloning}, an imitation learning framework 
    that redefines the action space using latent EP parameters ($K^{\mathrm{eq}}$, 
    $\theta^{\mathrm{eq}}$) extracted via a particle filter from bilateral 
    teleoperation demonstrations.
    \item We show that this representation transfers across unseen surface geometries 
    and object shapes in contact-rich tasks (wiping, pick-and-place, grasping), 
    where trajectory-based baselines fail.
    \item We demonstrate that the proposed action space is architecture-agnostic, 
    improving torque tracking performance when combined with both Transformer-based 
    (ILBiT) and SSM-based (Mamba) generative models.
\end{itemize}
 
\section{RELATED WORK}
\subsection{Imitation Learning for Contact-Rich Manipulation}
 
As discussed in Section~\ref{sec:introduction}, conventional imitation learning
methods~\cite{zhao2023act, chi2023diffusion, fu2024mobilealoha} define their action
spaces over observable robot states and therefore lack intrinsic compliance under
environmental deviations.
Even if a constant impedance is heuristically applied to a trajectory-generation model such as ACT, it theoretically struggles to fully adapt to unknown geometric variations because it lacks the intrinsic mechanism to dynamically infer and adjust the stiffness over time.
Bilateral control has been used as a data collection interface for imitation learning
to capture both position and force information simultaneously~\cite{adachi2018bilateral,
sasagawa2020bilateral}, though its role has remained limited to recording demonstrations.
While Bogdanovic et al.~\cite{bogdanovic2020variable} demonstrated that using impedance
parameters as actions improves robustness in reinforcement learning, our work extends
this insight to imitation learning, grounding the action space in the EP hypothesis.
 
\subsection{Variable Impedance Control and Stiffness Adaptation}
 
Data-driven stiffness adaptation has gained traction for contact-rich
tasks~\cite{ren2018compliance, beltran2020peg, oikawa2021assembly}.
Early work by Buchli et al.~\cite{buchli2011variable} demonstrated variable impedance
control via reinforcement learning (PI$^2$), establishing the principle that
impedance parameters can be treated as learnable control outputs. Okada et al.~\cite{okada2024contact} learn contact dynamics with a diffusion model and select
optimal stiffness offline via Bayesian optimization. Geiger et al.~\cite{geiger2026diffusion} reconstruct a contact-consistent simulated zero-force trajectory using a diffusion model conditioned on measured wrenches to implicitly adapt robot stiffness. Another line of research estimates stiffness directly from human demonstrations by mapping the positional variance across multiple demonstrations to stiffness profiles, assuming that regions with low variance require high stiffness~\cite{zhang2021learning}. While effective, this heuristic approach inherently requires multiple demonstrations of the exact same task to compute statistical variance, and does not capture the true physical compliance of the human arm. In contrast, our method takes the \emph{human operator's} stiffness, together with an equilibrium point estimated as a quantity distinct from the measured configuration, as the direct targets for imitation learning. Because human stiffness and equilibrium points encode the adaptive strategy used to handle uncertainty, they can be estimated from a single demonstration and transfer across environmental conditions without re-identification.
 
\subsection{The Equilibrium Point Hypothesis in Motor Control and Robotics}
 
The EP hypothesis~\cite{feldman1966functional, feldman1986, bizzi1992does} proposes that 
the central nervous system regulates movement via a virtual equilibrium 
configuration $\theta^{\mathrm{eq}}$ and stiffness 
$K^{\mathrm{eq}}$~\cite{burdet2001central}. 
The same spring-like abstraction underlies impedance control in
robotics~\cite{hogan1985impedance}, where the equilibrium configuration and
stiffness are prescribed as controller settings rather than learned.
Tele-impedance~\cite{ajoudani2012tele, peternel2023decade} transfers the
operator's arm stiffness, typically estimated from EMG signals, to a remote
manipulator in real time, and such human-in-the-loop demonstrations have also
been used to teach robots motion and stiffness profiles for autonomous
execution~\cite{peternel2014teaching}.
These approaches, however, rely on dedicated sensing of the operator and take
the measured hand position as the motion reference; the equilibrium point is
not estimated as a quantity distinct from the actual configuration.
To our knowledge, no prior work has treated EP parameters as a
\emph{target representation} for imitation learning.
Hou et al.~\cite{hou2025acp} note that recovering full compliance parameters
from human demonstrations is ill-posed, and therefore construct an approximate
compliance profile by permitting compliance only along the measured force
direction.
In contrast, our method explicitly takes the operator's actual stiffness and equilibrium point
as the imitation-learning targets, demonstrating that these can be estimated from a single
demonstration using only end-effector position and force, or joint angle and torque. We extract these parameters as latent time
series from teleoperation data and reproduce them, either by direct replay or with a trained policy, enabling robust
generalization to unseen environmental conditions.

\section{PROPOSED METHOD}
\label{sec:method}
\begin{figure}[t]
        \centering
        \includegraphics[width=\columnwidth]{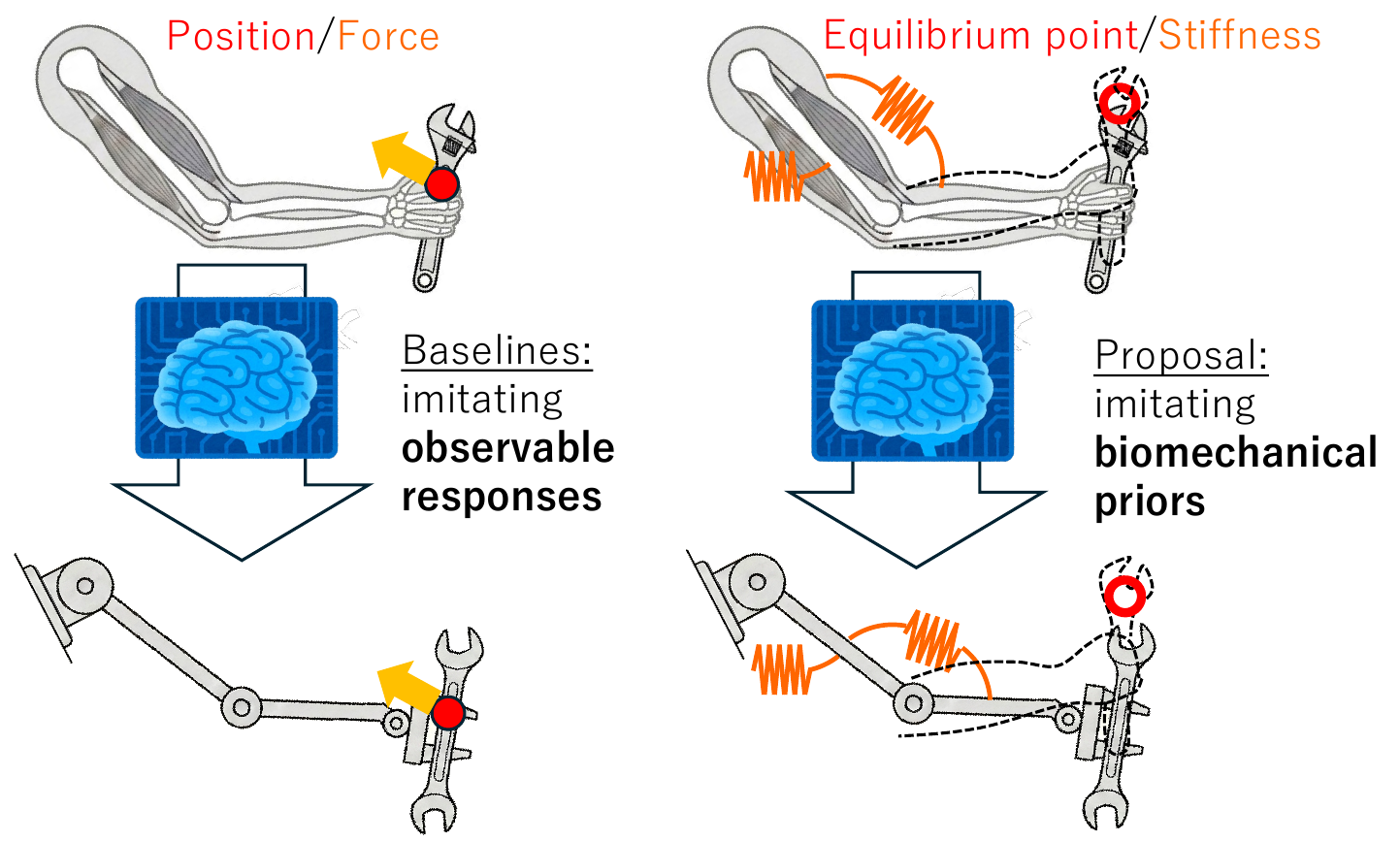}
        \caption{Comparison of imitation targets. Baselines imitate observable responses (position and force); the proposed method imitates the latent parameters that generate them---equilibrium point and stiffness---enabling passive compliance against environmental uncertainty.}
        \label{fig:system_pipeline}
    \end{figure}

    We refer to the proposed framework as \textit{Impedance Cloning}, 
    whose architecture comprises parameter estimation, imitation learning, 
    and torque command generation. The key conceptual distinction from 
    conventional methods is illustrated in Fig.~\ref{fig:system_pipeline}.
    
    Bilateral teleoperation plays a specific role in this construction. The operator's
    command is not directly observable. In a bilateral system, the human arm and the leader
    robot share a single mechanical state. The joint angle and the torque estimated by the
    reaction force observer are therefore acquired from the very leader robot to which the
    human applies force. The EP model of
    \eqref{eq:ep_model} below is the assumption that converts this observable pair into
    the latent command. The leader robot is therefore used twice: as the measuring
    instrument during demonstration, and as the executor of the reconstructed command
    during autonomous operation.
    
\subsection{Definition of a Mathematical Model}
    Based on the EP hypothesis, the operator's command is modeled as a spring acting between a virtual equilibrium configuration and the measured configuration of the leader robot. Let $\xi$ denote the measured configuration and $f$ the generalized force conjugate to it. The model is then written as
    
    \begin{equation}
        f = K^{\mathrm{eq}}(\xi^{\mathrm{eq}} - \xi).
        \label{eq:ep_model}
    \end{equation}
    
    Here, $K^{\mathrm{eq}}$ is the stiffness and $\xi^{\mathrm{eq}}$ is the equilibrium point. We instantiate \eqref{eq:ep_model} in two coordinate frames. In joint space, the human operational torque $\tau$ is given by
    
    \begin{equation}
        \tau = K^{\mathrm{eq}}(\theta^{\mathrm{eq}} - \theta),
        \label{eq:ep_model_joint}
    \end{equation}
    
    where $\theta$ is the leader robot's joint angle measured during teleoperation and $\tau$ is the joint torque estimated by the reaction force observer. In task space, the human operational force $F$ is given by
    
    \begin{equation}
        F = K^{\mathrm{eq}}(x^{\mathrm{eq}} - x),
        \label{eq:ep_model_task}
    \end{equation}
    
    where $x$ is the leader's end-effector position obtained by forward kinematics and $F = J^{-T}\tau$ is the corresponding operational force, with $J$ the Jacobian of the leader robot.
    
    In both cases, this work considers only the stiffness of each degree of freedom and does not account for the off-diagonal components of the stiffness matrix; an independent estimator is therefore assigned to each joint or each translational axis. Because $\xi$ is defined in the leader's coordinate frame, $K^{\mathrm{eq}}$ and $\xi^{\mathrm{eq}}$ represent the operator's command parameters projected onto that frame rather than muscle-level quantities. The choice of the frame is therefore a design choice of the action space, not a restriction of the model. In the proposed method, the state variable $z_t = [K^{\mathrm{eq}}_t, \xi^{\mathrm{eq}}_t]^T$ based on this model is estimated offline at each time step using a particle filter and is used as the target for subsequent learning.
 
\subsection{Parameter Estimation Algorithm}
    For readability, the estimator is described below in the joint-space notation of \eqref{eq:ep_model_joint}; substituting $(x, F)$ for $(\theta, \tau)$ yields the task-space estimator of \eqref{eq:ep_model_task} without any other modification.
    At a single instant, \eqref{eq:ep_model_joint} provides one equation for two unknowns, so estimating $(K^{\mathrm{eq}}, \theta^{\mathrm{eq}})$ from joint angle and torque data alone is ill-posed. What makes it resolvable is the content of the EP hypothesis itself: the assumption that the command parameters vary slowly compared with the measured configuration $\theta$. Under this assumption, responses observed at different joint angles within a short window jointly constrain the pair, and sequential Bayesian estimation recovers it. Because this work treats stiffness as a per-degree-of-freedom scalar, we employ an independent particle filter for each joint. 
    In the state transition model, the parameters primarily undergo a minor random walk. Furthermore, to prevent the estimated values from diverging in steady states, an adaptive micro-transition process depending on the magnitude of the stiffness is incorporated into the state transition process. Specifically, a nonlinear state transition equation is applied, which pulls $\theta^{\mathrm{eq}}$ toward the current joint angle $\theta$ when stiffness is small, and attenuates this effect when the stiffness is large. The state transition at time $t$ for the $i$-th particle is formulated as follows:
    
    \begin{equation}
        \theta_{t,i}^{\mathrm{eq}} = \theta_{t-1,i}^{\mathrm{eq}}
            + \alpha_{t,i} (\theta_t - \theta_{t-1,i}^{\mathrm{eq}})
            + \epsilon^{\theta}_{t,i},
        \label{eq:transition_theta}
    \end{equation}
    \begin{equation}
        K_{t,i}^{\mathrm{eq}} = K_{t-1,i}^{\mathrm{eq}} + \epsilon^{K}_{t,i}.
        \label{eq:transition_k}
    \end{equation}
 
    Here, $\alpha_{t,i}$ is a pull-back coefficient that adaptively changes according to the magnitude of the stiffness, defined by the following equation:
 
    \begin{equation}
        \alpha_{t,i} = \frac{\gamma}{1 + \beta K_{t-1,i}^{\mathrm{eq}}},
        \label{eq:pullback}
    \end{equation}
 
    where $\gamma$ and $\beta$ are hyperparameters controlling the strength of the pull-back. Furthermore, $\epsilon^\theta_{t,i} \sim \mathcal{N}(0, \sigma_\theta^2)$ and $\epsilon^K_{t,i} \sim \mathcal{N}(0, \sigma_K^2)$ represent system noise for the random walk of each parameter. To ensure the physical consistency of the stiffness, a lower bound is applied to $K_{t,i}^{\mathrm{eq}}$ after every update to prevent negative values.
    
    In the observation model, the squared error between the measured torque at time $t$ and the estimated torque calculated using the mathematical model from the estimated states $K_{t,i}^{\mathrm{eq}},\theta_{t,i}^{\mathrm{eq}}$ held by the $i$-th particle and the measured angle is evaluated to calculate the likelihood $L_{t,i}$.
    
    \begin{equation}
        L_{t,i} = \exp \left( -\frac{\left( K^{\mathrm{eq}}_{t,i}
            (\theta^{\mathrm{eq}}_{t,i} - \theta_t) - \tau_t \right)^2}
            {2\sigma^2} \right).
        \label{eq:likelihood}
    \end{equation}
    
    Here, $\sigma$ is the standard deviation of the observation noise. Resampling is performed based on this likelihood, and a weighted average is calculated to stably estimate the parameters at each time step.

All hyperparameters are listed in Table~\ref{tab:pf}; they were determined
empirically to balance the motion stability of the physical robot and its
responsiveness to environmental changes.
 
\begin{table}[t]
  \centering
  \caption{Particle filter hyperparameters. Because particles are truncated
  during resampling, the effective particle count is lower than the nominal
  $N_p$, ranging from 526 to 959 across trials.}
  \label{tab:pf}
  \begin{tabular}{lll}
    \toprule
    Parameter & Symbol & Value \\
    \midrule
    Number of particles       & $N_p$        & 1000  \\
    Sampling period           & $\Delta t$   & 2\,ms \\
    Joint angle noise std.    & $\sigma_\theta$ & 0.005\,rad \\
    Stiffness noise std.      & $\sigma_K$   & 0.02\,Nm/rad \\
    Observation noise std.    & $\sigma$     & 0.1\,Nm \\
    Pull-back base strength   & $\gamma$     & 0.02  \\
    Pull-back decay rate      & $\beta$      & 5.0   \\
    Likelihood threshold      & ---          & 0.8   \\
    Resampling interval       & ---          & every 10 steps \\
    Stiffness lower bound     & ---          & 0.01\,Nm/rad \\
    \bottomrule
  \end{tabular}
\end{table}
 
\subsection{Implementation of the Imitation Learning Module}
\begin{figure}[t]
            \centering
            \begin{subfigure}[b]{0.38\columnwidth}
                \centering
                \includegraphics[width=\linewidth]{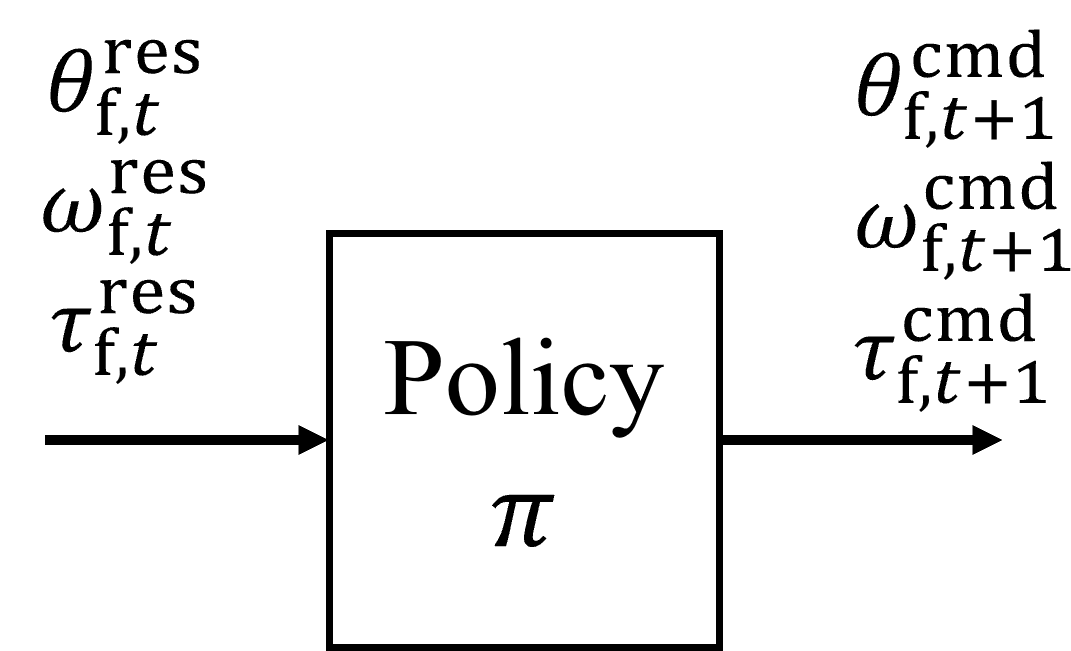}
                \caption{Baseline Method}
                \label{fig:action_space_conv}
            \end{subfigure}
            \hfill
            \begin{subfigure}[b]{0.59\columnwidth}
                \centering
                \includegraphics[width=\linewidth]{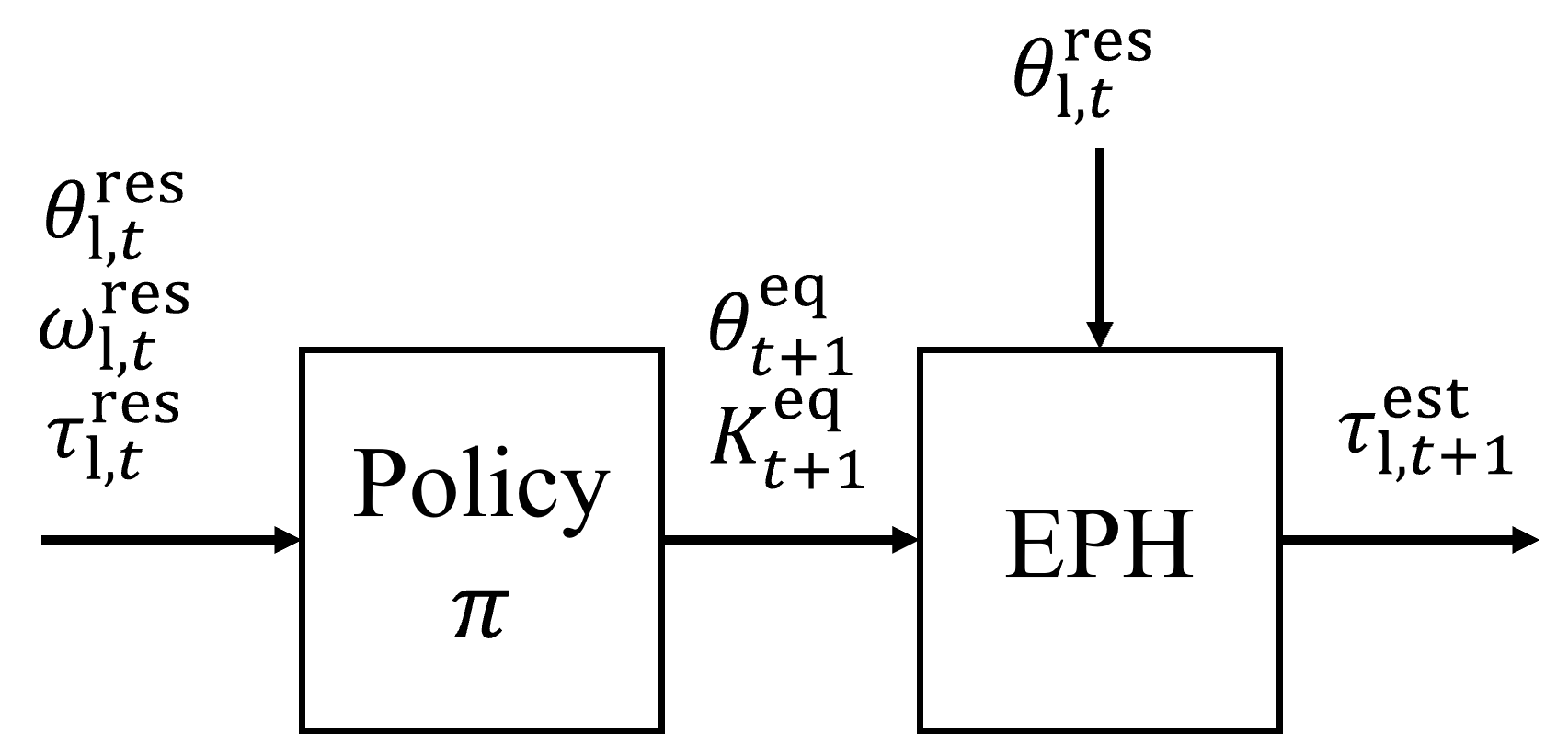}
                \caption{Proposed Method}
                \label{fig:action_space_prop}
            \end{subfigure}
            
            \caption{Comparison of action spaces.
(a) The baseline method directly outputs spatial trajectories such as target joint angles and torques.
(b) The proposed method defines the action space using upper-level physical parameters (stiffness $K^{\mathrm{eq}}$ and equilibrium point $\theta^{\mathrm{eq}}$), enabling compliance and robust adaptation to environmental variations.}
            \label{fig:action_space}
        \end{figure}

    Fig.~\ref{fig:action_space} compares the action spaces of the baseline and proposed approaches.
    Conventional imitation learning methods represent robot actions as $a_t = [\theta_t, \omega_t, \tau_t]$.
    In contrast, the proposed method represents actions as impedance parameters $a_t = [\theta^{\mathrm{eq}}_t, K^{\mathrm{eq}}_t]$.
    For each backbone, both methods share the same execution scheme---direct replay for Motion Copy, and the same architecture and training procedure for the learned models---differing only in the action representation.
    
    In this study, we evaluated configurations across different task settings.
    For the wiping and pick-and-place experiments, we employ \emph{Motion Copy}~\cite{yokokura2009motioncopy} as the backbone. Motion Copy essentially functions as a fixed-impedance control by directly replaying the recorded position, velocity, and force action sequences from demonstrations without a generative model.
    In the pick-and-place task, we additionally introduce a Variable Impedance Control (VIC) baseline based on the variance-based stiffness estimation method~\cite{zhang2021learning}, which derives stiffness inversely proportional to the positional variance across a set of demonstrations. We also evaluate the Motion Copy baseline under a single fixed-impedance setting. In the Cartesian-space wiping experiment, the Motion Copy baseline is instead evaluated under three fixed-impedance levels, obtained by scaling the position gain as $K_{\mathrm{p}}$, $0.75K_{\mathrm{p}}$, and $0.5K_{\mathrm{p}}$.
    \emph{ILBiT} and \emph{Mamba} are used as generative sequence modeling backbones in the grasping experiment to assess adaptability across different sequence modeling architectures.
    
    This framework allows us to assess both data-efficient replay-based performance and the generalization capability of generative models under the proposed and conventional action representations.

\subsection{Formulation of Torque Command Generation Rules}
\begin{figure}[t]
        \centering
        \begin{subfigure}[b]{\columnwidth}
            \centering
            \includegraphics[width=0.8\linewidth]{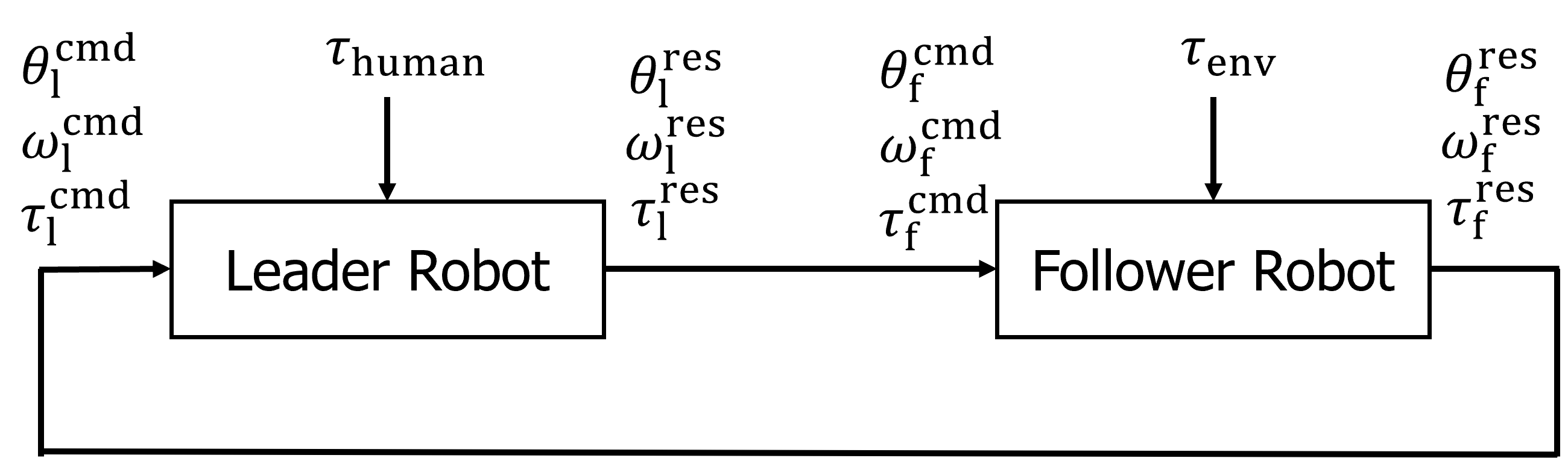}
            \caption{Macro Bilateral Architecture}
            \label{fig:control_macro}
        \end{subfigure}

        \vspace{2mm}

        \begin{subfigure}[b]{\columnwidth}
            \centering
            \includegraphics[width=\linewidth]{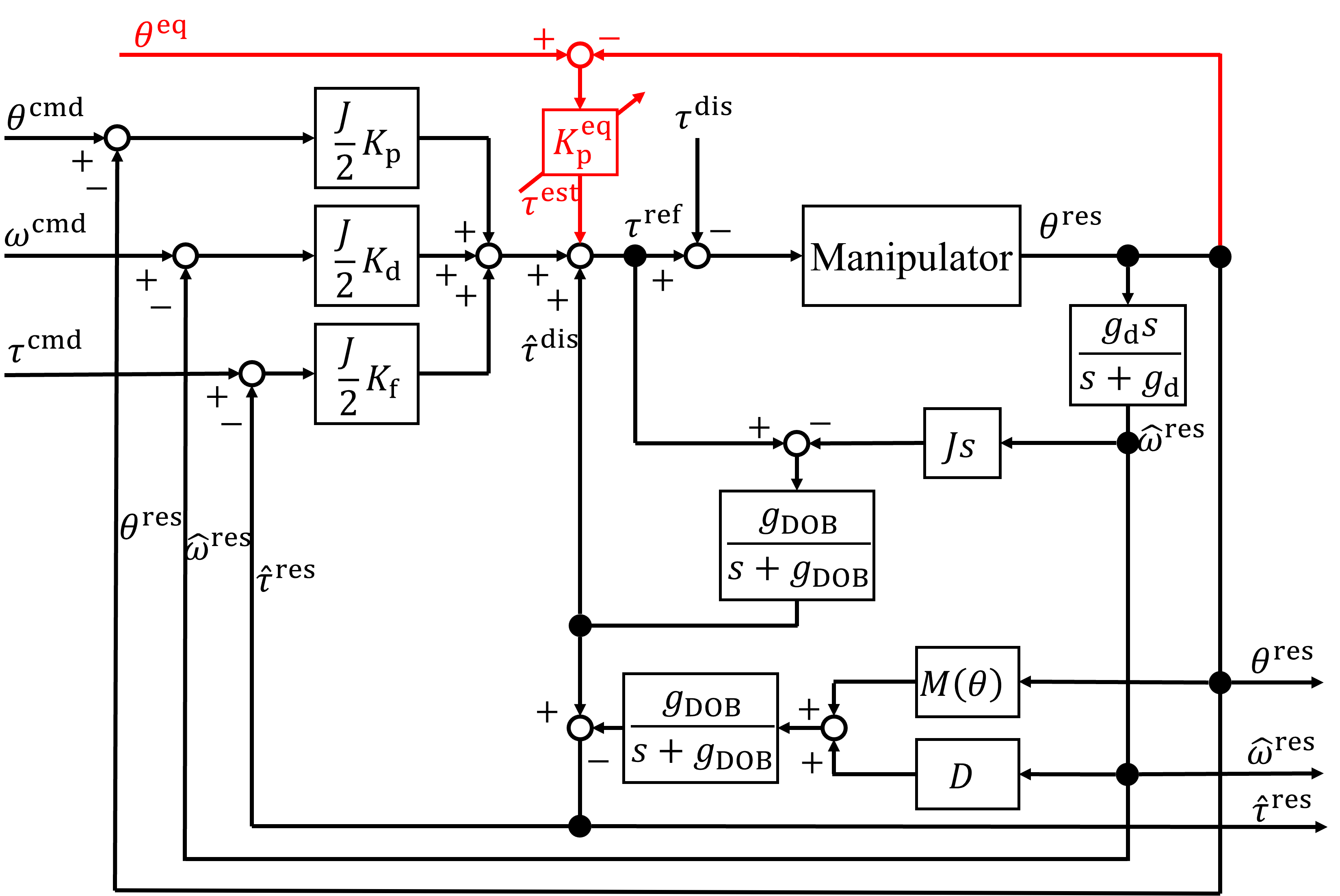}
            \caption{Leader Control}
            \label{fig:control_leader}
        \end{subfigure}
        \caption{Control architecture of the proposed method during autonomous execution.
(a) Signal flow between the leader and follower robots.
(b) Leader control. The red components compute the human operational torque from $K^{\mathrm{eq}}$ and $\theta^{\mathrm{eq}}$; the black loop alone is the standard feedback control used by the follower.}
        \label{fig:control_all}
    \end{figure}

    The control architecture during autonomous execution in the proposed method is shown in
    Fig.~\ref{fig:control_all}, following the disturbance-observer-based bilateral
    framework~\cite{ohnishi1996mechatronics}. In the macro bilateral architecture
    (Fig.~\ref{fig:control_all}(a)), the follower's responses ($\theta^{\mathrm{f}}, \omega^{\mathrm{f}}, \tau^{\mathrm{f}}$) serve as the leader's command references ($\theta^{\mathrm{l, cmd}}, \omega^{\mathrm{l, cmd}}, \tau^{\mathrm{l, cmd}}$). The final torque command value $\tau^{\mathrm{l, ref}}$ for the leader robot is formulated by the following equation using the moment of inertia $J$ of each joint, along with the position gain $K_{\mathrm{p}}$, velocity gain $K_{\mathrm{d}}$, and force gain $K_{\mathrm{f}}$:
    \begin{equation}
        \begin{split}
        \tau^{\mathrm{l, ref}} = {}& \tau^{\mathrm{human, est}}
          + \frac{J}{2} K_{\mathrm{p}} (\theta^{\mathrm{l, cmd}} - \theta^{\mathrm{l, res}}) \\
          &+ \frac{J}{2} K_{\mathrm{d}} (\omega^{\mathrm{l, cmd}} - \omega^{\mathrm{l, res}})
          + \frac{1}{2} K_{\mathrm{f}} (\tau^{\mathrm{l, cmd}} - \tau^{\mathrm{l, res}}).
        \end{split}
        \label{eq:leader_torque}
    \end{equation}
 
    Here, $\theta^{\mathrm{l, res}}, \omega^{\mathrm{l, res}}, \tau^{\mathrm{l, res}}$ are the response values on the leader side. The first term on the right side, $\tau^{\mathrm{human, est}}$, is the estimated human operational torque component, calculated by the following equation using the stiffness $K^{\mathrm{eq}}$ and equilibrium point $\theta^{\mathrm{eq}}$ inferred by the network (red components in Fig.~\ref{fig:control_all}(b)).
    \begin{equation}
        \tau^{\mathrm{human, est}} = K^{\mathrm{eq}}
            (\theta^{\mathrm{eq}} - \theta^{\mathrm{l, res}})
            - D^{\mathrm{eq}} \omega^{\mathrm{l, res}}.
        \label{eq:human_torque}
    \end{equation}
 
    Here, $D^{\mathrm{eq}}$ is a constant damping coefficient introduced only at execution time. During demonstration, the operator physically holds the leader robot, so the damping of the human arm is present in the hardware itself and need not be commanded. This is consistent with the $\lambda$-model~\cite{feldman1986}, in which damping is not an independent command variable but a quantity tied to stiffness ($D^{\mathrm{eq}} = \mu K^{\mathrm{eq}}$); the estimator therefore outputs stiffness only. During autonomous execution, however, the operator's hand is absent, and the unheld leader robot exhibits high-frequency vibration unless this physical damping is replaced. We therefore add a constant $D^{\mathrm{eq}}$ per joint (0.1, 0.02, 0.11, 0.03, 0.09, 0.03, 0.07, and 0.1\,Nm$\cdot$s/rad for J1--J8) purely as a stabilization term for the physical manipulator, rather than as an estimate of the operator's damping. Omitting damping during estimation introduces negligible model mismatch: in the pick-and-place demonstration data, the ratio $D^{\mathrm{eq}}\omega / K^{\mathrm{eq}}\Delta\theta$ remains within an RMS of 0.7--10\% at the three load-bearing joints (J2, J4, and J8).
    
    The follower runs only the black loop in Fig.~\ref{fig:control_all}(b), so the inferred human operational torque enters the leader alone, forming an asymmetric configuration. As a result, intervention from the upper-level parameters is added to the feedback commands based on the reaction force from the follower, achieving adaptive and stable contact motion against environmental uncertainty. In the baseline, the leader is not used; the follower runs the same loop with the replayed or predicted $[\theta_t, \omega_t, \tau_t]$ as its command references.
    
\section{EXPERIMENTS}
\label{sec:experiments}
    We first confirm the validity of the estimated parameters (Experiment 1), and then validate the proposed method through three manipulation tasks: a wiping task for robustness against height perturbations (Experiment 2), a pick-and-place task for generalizability across objects (Experiment 3), and a grasping task to evaluate adaptability across generative sequence modeling architectures (Experiment 4).
\subsection{Test Environment and System Configuration}
\begin{figure}[t]
            \centering
            \includegraphics[width=0.7\columnwidth]{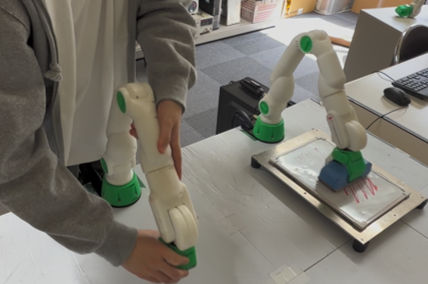}
            \caption{Experimental setup using two CRANE-X7 manipulators in a bilateral configuration. One acts as the leader operated by a human and the other as the follower.}
            \label{fig:setup}
        \end{figure}

    For the experiments, we constructed a bilateral control system using two 8-DOF CRANE-X7 manipulators (Fig.~\ref{fig:setup}). Joint torques were estimated using a reaction force observer without physical force/torque sensors. A commercially available whiteboard cleaner was grasped by the end-effector hand, and the target whiteboard was fixed to the workbench with tape.
 
\subsection{Dataset and Training Settings}
    For the wiping task, the operator wiped the cleaner back and forth three times to erase writing on the board via bilateral control. The operator was instructed to keep the motion cycle constant, and a time-series dataset totaling approximately 75,000 steps was constructed by acquiring 10 samples of motion at a sampling period of 2\,ms. For the pick-and-place task, the operator demonstrated picking and placing a single cup (Type A in Fig.~\ref{fig:containers}) from a fixed source position to a fixed target position, collecting 10 demonstrations under the same bilateral control setup. Both the proposed method and the Motion Copy baseline---which operates under fixed-impedance control---are executed using only one of these ten demonstrations. In contrast, the VIC baseline uses all ten demonstrations to calculate the variance for stiffness derivation. For the grasping task, the operator demonstrated grasping a single cup stably, collecting 16 demonstrations for training. To perform learning independent of robot-specific characteristics, we used only unique time-series data acquired from the physical robot. Because the proposed method targets physical upper-level parameters for learning, it exhibits sufficient generalization performance even with these small datasets.
    
    Because this work evaluates the effect of the action representation, the comparison must be made under a common input modality. ACT and Diffusion Policy are both designed to generate actions from camera observations, so comparing them directly with our framework, which uses no visual input, would leave it unclear whether a performance difference stems from the action representation or from the input modality. We therefore adopt ILBiT and Mamba, which share the same torque-based input structure as the proposed method, as the comparison targets.
 
ILBiT~\cite{kobayashi2024ilbit} is a Transformer-based~\cite{vaswani2017attention}
imitation learning model designed for bilateral control systems that takes joint
angle and torque sequences as input.
We refer the reader to the original paper for architectural details
and use the same hyperparameters reported therein.
 
Mamba~\cite{gu2023mamba,tsuji2025mamba} is a State Space Model (SSM)-based architecture
that compresses past context into low-dimensional state variables,
enabling real-time motion generation from limited training data.
The model consists of a single Mamba block with SSM dimensionality 4,
one input and three output linear layers ($n_i=1$, $n=3$, $n_o=1$),
and dropout with retention probability 0.75.
The matrix $A$ is fixed rather than learned, with diagonal elements
set to $\mathrm{diag}[-0.5, -0.2, -0.08, -0.032]$, following the
ablation results in the original paper.
Both models were trained with MSE loss and the AdamW optimizer
(learning rate $10^{-3}$), with a control loop cycle of 2\,ms and
inference communicated at a 100\,ms cycle via UDP.
 
\subsection{Evaluation Criteria and Experimental Conditions}
    For the \textbf{wiping task}, the primary metric is the average contact force $F_{\mathrm{avg}}$ during wiping motion, calculated from the vertical force $F_z(t)$ between the contact start time $t_{\mathrm{s}}$ and end time $t_{\mathrm{f}}$:
    \begin{equation}
        F_{\mathrm{avg}} = \frac{\Delta t}{t_{\mathrm{f}} - t_{\mathrm{s}}}
            \sum_{t=t_{\mathrm{s}}}^{t_{\mathrm{f}}} F_z(t).
        \label{eq:favg}
    \end{equation}
    In joint space, the board height was varied from $-10$ to $+10$\,cm in increments of 2\,cm from the reference position used in the demonstrations, with $\pm 1$\,cm conditions added to confirm finer variations. In Cartesian space, the range was $-8$ to $+8$\,cm (4\,cm steps for $0.5K_{\mathrm{p}}$ and $0.75K_{\mathrm{p}}$). Five trials were conducted under each condition.
 
    For the \textbf{pick-and-place task}, the primary metric is the task success rate across 10 cup types of varying shape and size. A trial is considered successful when the cup is transported from the source position to the target position without being dropped. For each method, 10 trials were conducted per cup type, resulting in a total of 100 trials.

Torque tracking error is measured as the root mean squared error (RMSE)
after Dynamic Time Warping (DTW) alignment between the evaluated
trajectory and a reference trajectory derived from training data.
The reference is constructed by first time-normalizing all training
trajectories to $N = 500$ points via linear interpolation, then
averaging them across trials.
Because the sign convention of the measured torque in the follower
differs from that of the leader, the reference values are sign-flipped
before comparison.
 
Given a test trajectory $s$ and the reference $r$, both of length
$N$, DTW computes an optimal warping path
$\mathcal{P} = \{(i_l, j_l)\}_{l=1}^{L}$ that minimizes the
cumulative distance between aligned samples.
The DTW-RMSE for a single joint is then defined as
\begin{equation}
  \mathrm{RMSE} = \sqrt{\frac{1}{L}\sum_{l=1}^{L}
    \bigl(s_{j_l} - r_{i_l}\bigr)^2}.
  \label{eq:dtw_rmse}
\end{equation}
\subsection{Experiment 1: Validity of Parameter Estimation}
    Under the settings described in Section~\ref{sec:method} (Table~\ref{tab:pf}), we verified
    the estimator using demonstration data from the pick-and-place task. The two-parameter model
    reconstructs the measured torque with a normalized RMSE (NRMSE) of 0.071.
 
    We conducted a sensitivity analysis to verify that the sequential estimation stably resolves
    the inherently ill-posed parameter estimation. Varying the number of particles $N_p$, the
    stiffness noise standard deviation $\sigma_K$, the initial values, and the random seeds gave a
    spread in the estimated $K^{\mathrm{eq}}$ ranging from 14\% (J3) to 40\% (J7) across joints.
    We further swept the pull-back hyperparameters over $\gamma \in [0.002, 0.1]$ and
    $\beta \in [0.2, 50]$. Excessive pull-back drives $\theta^{\mathrm{eq}}$ to converge
    prematurely to the current joint angle $\theta$, degrading the NRMSE of torque reconstruction
    from 0.071 to 0.275. Taking, for J1, J3 and J7, the ratio between the maximum and minimum mean
    $K^{\mathrm{eq}}$ over the $\gamma$--$\beta$ grid (49 settings, seed-averaged), this ratio
    ranged from 4.5 (J1) to 5.9 (J7). Restricting the sweep to hyperparameters whose
    reconstruction error lies within 5\% of the best value, however, keeps the ratio within a
    factor of 1.6 for all of J1, J3 and J7; that is, the estimates are stable once the
    hyperparameters are confined to the region where the torque is adequately reconstructed, and
    the values used in this work (Table~\ref{tab:pf}) fall inside this band.
 
\subsection{Experiment 2: Wiping Task --- Robustness to Height Perturbation}
\label{subsec:exp1}
    The height offset is absent from the demonstration data, so a trajectory-level action representation has no
    information with which to respond to it, whereas an EP-level representation carries the operator's own
    stiffness. We therefore expect a much smaller force–height slope for the proposed method, and a
    slope that becomes steeper as the position gain increases for fixed-impedance replay.
    
    This experiment was conducted under two conditions: with the action space defined in joint
    space~\eqref{eq:ep_model_joint} and in Cartesian space~\eqref{eq:ep_model_task}. The
    baseline was a single fixed-impedance setting in joint space, and fixed-impedance settings at
    three position-gain levels in Cartesian space.
    
\begin{figure}[t]
        \centering
        \includegraphics[width=0.98\columnwidth]{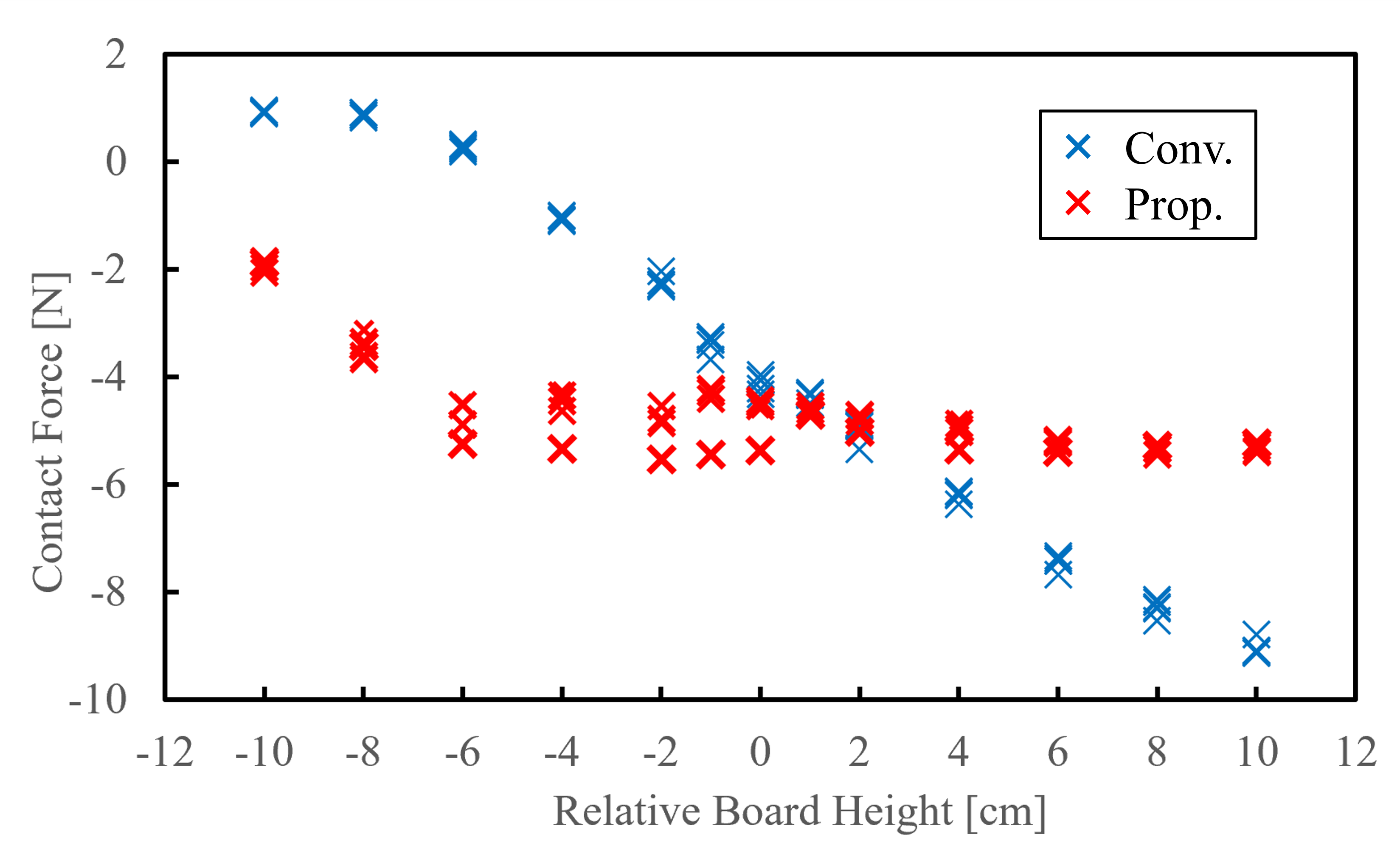}
        \caption{Average wiping contact force versus board height offset in joint space (each marker: one trial). Negative values indicate pressing force.}
        \label{fig:force_result}
    \end{figure}

    In the joint-space condition, as shown in Fig.~\ref{fig:force_result}, the contact force of
    the baseline fluctuated significantly in proportion to the change in board height.
    Particularly when the height decreased, the robot failed to maintain contact because the
    recorded trajectories were rigidly replayed; under conditions of $-6$\,cm or lower, the
    contact force almost disappeared. In contrast, the proposed method maintained the desired
    contact force of approximately 4--5\,N within the range of $-6$ to $+10$\,cm, regardless of
    the environmental height change. The inferred stiffness $K^{\mathrm{eq}}$ and equilibrium
    point $\theta^{\mathrm{eq}}$ provide compliance that naturally absorbs unknown positional
    deviations to maintain stable contact forces. Below $-6$\,cm, however, the contact force of
    the proposed method also decreased gradually. This is attributable to defining the action
    space in joint space: translational errors are absorbed through joint rotations, which causes
    unintended posture fluctuations, and the tilted end-effector makes the edge of the cleaner
    physically catch on the board.
    
\begin{figure}[t]
        \centering
        \includegraphics[width=0.95\columnwidth]{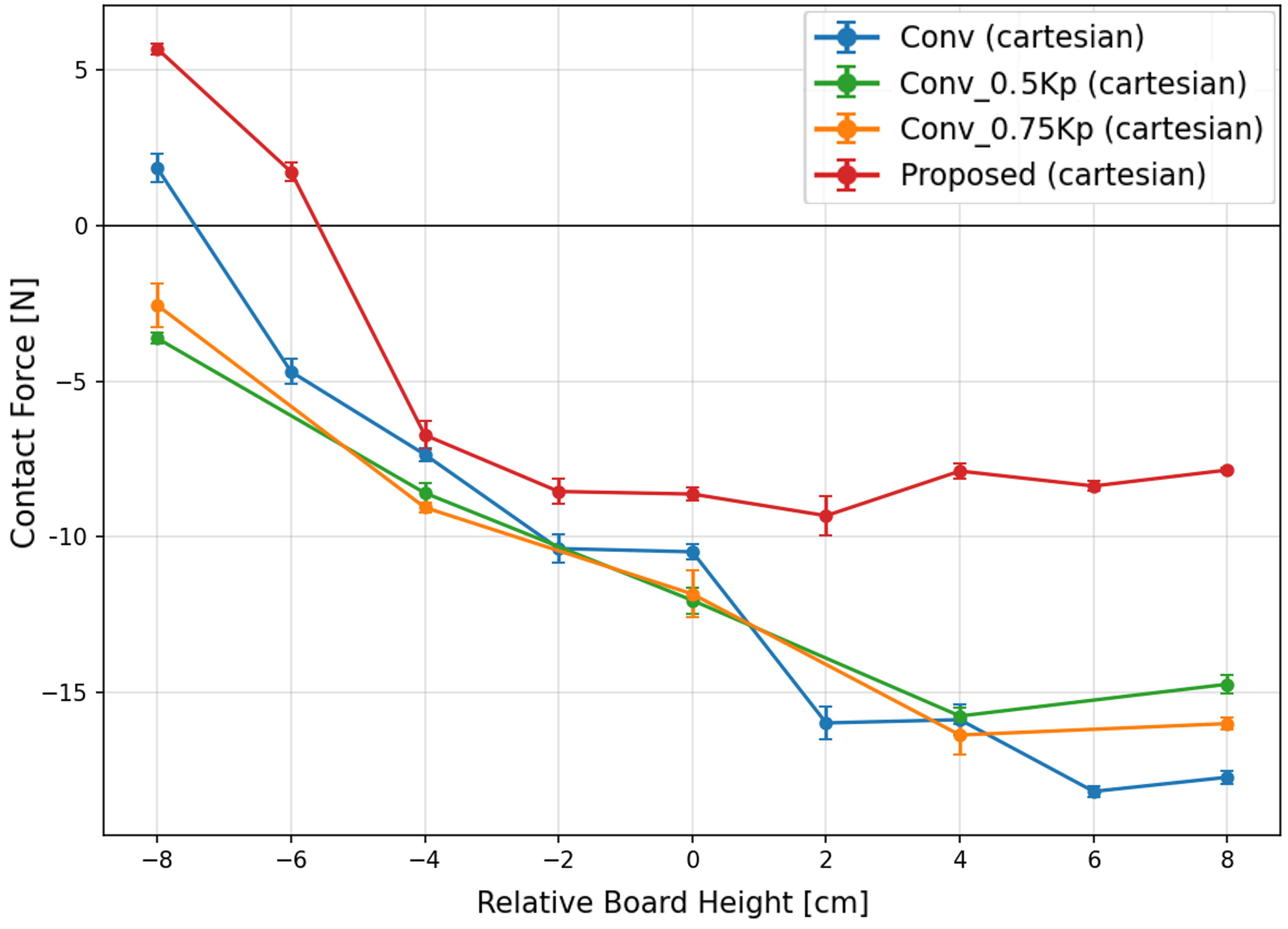}
        \caption{Average wiping contact force versus board height offset in Cartesian space (mean $\pm$ SD over five trials). Negative values indicate pressing force.}
        \label{fig:cartesian_result}
    \end{figure}

    In the Cartesian-space condition, the action space was parameterized by Cartesian position and
    force instead of joint angle and torque. As shown in Fig.~\ref{fig:cartesian_result}, the
    proposed method exhibited the same qualitative robustness to height offsets, achieving a force–height slope of $0.13 \pm 0.03$\,N/cm over 0 to $+8$\,cm (absolute value; mean $\pm$ SD of per-trial linear fits). In contrast, the three fixed-impedance baselines exhibited steeper slopes of $0.34 \pm 0.06$, $0.52 \pm 0.09$, and $0.83 \pm 0.05$\,N/cm for $0.5K_{\mathrm{p}}$,
    $0.75K_{\mathrm{p}}$, and $K_{\mathrm{p}}$, respectively, indicating that sensitivity to
    positional deviations grows monotonically with the position gain.
    At offsets of $-6$\,cm and below, the proposed method lost contact, likely because wiping beyond the equilibrium point reverses the restoring force in~\eqref{eq:ep_model_task} away from the board; the abrupt change between $-4$ and $-6$\,cm supports this interpretation.
    
    These results match the pattern predicted above and confirm that Impedance Cloning is
    effective and applicable in either coordinate space.
    
\subsection{Experiment 3: Pick-and-Place Task --- Generalizability Across Object Types}

\begin{figure}[t]
  \centering
  \includegraphics[width=0.65\linewidth]{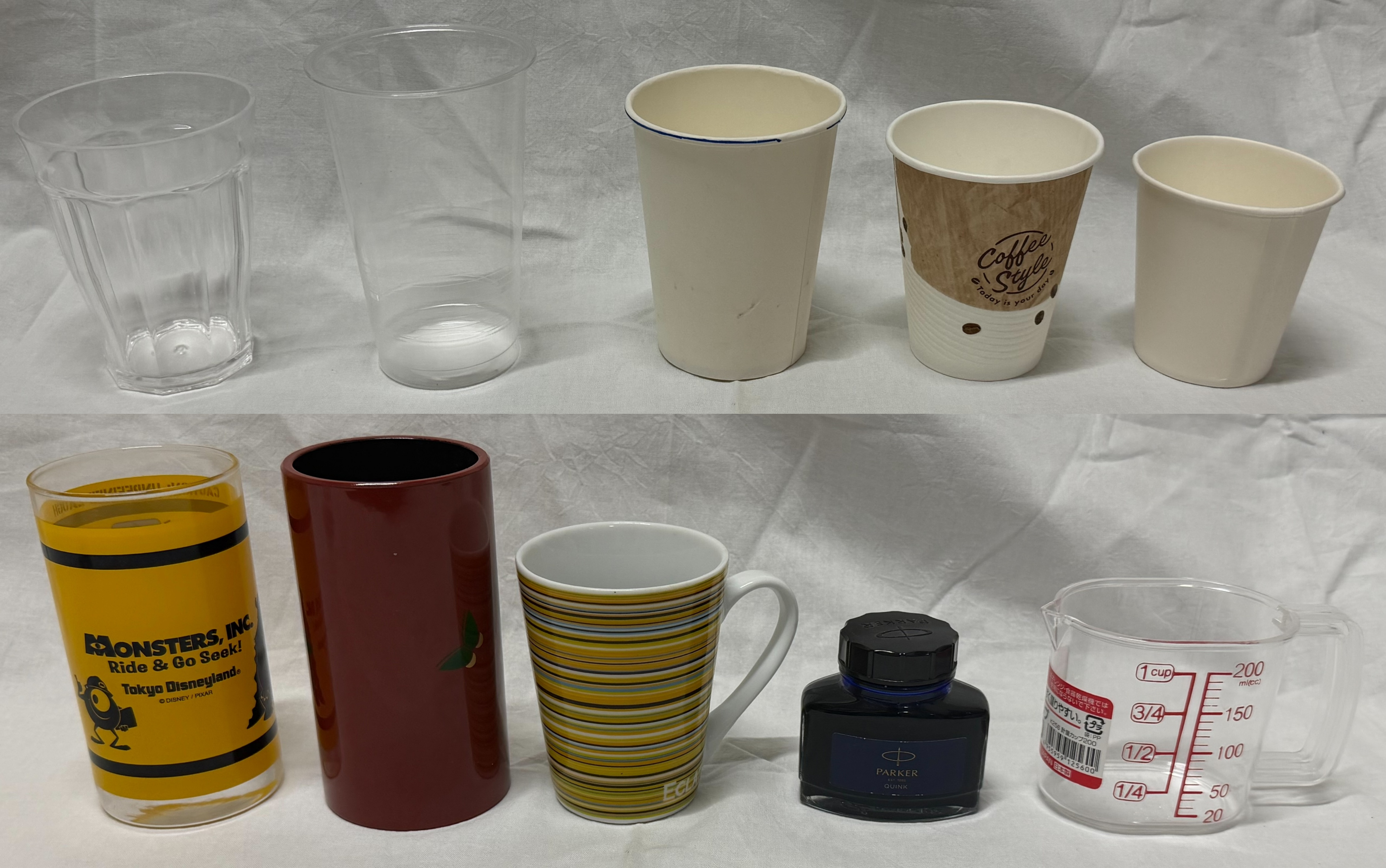}
  \caption{The 10 test containers used for evaluation.}
  \label{fig:containers}
\end{figure}

Fig.~\ref{fig:containers} shows the 10 evaluation containers.
They span a range of materials (paper, plastic, glass, rubber-coated),
diameters (approximately 5--10\,cm), and heights (approximately
7--15\,cm), providing varied combinations of surface stiffness,
contact geometry, and grasp compliance.
All containers were placed at the same fixed position as the
demonstration cup; no positional perturbation was introduced so that
the only source of variation was the object's physical properties.
 
    Table~\ref{tab:pick_place} summarizes the success rates of the pick-and-place task involving 10 cup types of varying shapes and materials, 9 of which were not used in the demonstration.
    
    \begin{table}[t]
        \centering
        \caption{Pick-and-place task success results for 10 cup types. VIC: variable impedance control baseline; FI: fixed-impedance (Motion Copy) baseline.}
        \label{tab:pick_place}
        \begin{tabular}{lccc}
            \toprule
            Cup Type & Proposed & VIC & FI \\
            \midrule
            Hard Plastic Cup (Type A) & 6 / 10 & 5 / 10  & 10 / 10 \\
            Soft Plastic Cup          & 10 / 10 & 8 / 10  & 10 / 10 \\
            Large Paper Cup           & 10 / 10 & 10 / 10 & 10 / 10 \\
            Medium Paper Cup          & 10 / 10 & 9 / 10  & 5 / 10  \\
            Small Paper Cup           & 10 / 10 & 10 / 10 & 3 / 10  \\
            Glass Cup                 & 10 / 10 & 10 / 10 & 10 / 10 \\
            Tall Ceramic Cup          & 7 / 10  & 10 / 10 & 10 / 10 \\
            Ceramic Mug               & 8 / 10  & 7 / 10  & 10 / 10 \\
            Ink Bottle                & 4 / 10  & 7 / 10  & 0 / 10  \\
            Measuring Cup             & 9 / 10  & 6 / 10  & 6 / 10  \\
            \midrule
            Total & 84 / 100 & 82 / 100 & 74 / 100 \\
            \bottomrule
        \end{tabular}
    \end{table}
 
    The proposed method achieved a total success rate of 84/100, outperforming the fixed-impedance baseline, which scored 74/100.
    Furthermore, despite using only a single demonstration, the proposed method performed comparably to the VIC baseline (82/100), with no significant difference observed (Fisher's exact test).
    The improvement of the proposed method over the fixed-impedance baseline strongly correlated with the containers' deviation from the required fixed-impedance parameters (Spearman $\rho=0.88, p=0.0008, n=10$).
    Interestingly, the largest single loss for the proposed method occurred on the demonstration cup (Type A, $-4/10$ compared to the baseline), while all containers whose gripper joint angle differed from the demonstration cup by more than 0.40\,rad saw improvements of $+3$ to $+7/10$. This indicates that operating in the EP parameter space robustly adapts to geometric deviations through compliance, whereas fixed trajectories struggle to accommodate such variations.
 
\subsection{Experiment 4: Grasping Task}
 
We evaluated the proposed method on a top-down grasping task with 16 demonstrations on a single cup as training data and 10 cups of varying stiffness and geometry as test objects.
To isolate the effect of Impedance Cloning, we adopted ILBiT~\cite{kobayashi2024ilbit} and
Mamba~\cite{tsuji2025mamba} as representative Transformer-based and
SSM-based architectures that operate purely on position and force signals.
For each backbone, the baseline variant directly imitates position and
force trajectories, while the proposed variant replaces the action space
with the equilibrium point parameters $K^{\mathrm{eq}}$ and
$\theta^{\mathrm{eq}}$.
 
The results are summarized in Table~\ref{tab:results}; the reported values are averaged over all 10 test containers and over the gripper joint alone or all 8 joints, respectively.
The baseline variants exhibited large and variable errors due to the inability of trajectory policies to adapt to unseen object properties. In contrast, the proposed method consistently reduced both mean error and variance across all containers and architectures. Note that the overlapping standard deviations in Table~\ref{tab:results} reflect across-container variability rather than across-trial variability; a per-container comparison under the same backbone gives Welch's t-test $p=0.02$ (ILBiT) and $p=0.002$ (Mamba) at the gripper joint. The compliance encoded in the EP parameters enables the robot to absorb object-level variations passively for stable force regulation.
 
\begin{table}[t]
  \centering
  \caption{%
    Torque tracking error~[Nm] on 10 unseen containers (mean $\pm$ SD).
    Baseline: each architecture trained to directly imitate joint angle, angular velocity, and torque trajectories.
    Prop.: same architecture with the proposed equilibrium point action space.
  }
  \label{tab:results}
  \begin{tabular}{llcc}
    \toprule
    Architecture & Method & Gripper joint & All joints \\
    \midrule
    \multirow{2}{*}{ILBiT}
      & Baseline & $0.0831 \pm 0.0387$ & $0.0674 \pm 0.0794$ \\
      & Prop.    & $\mathbf{0.0477 \pm 0.0093}$ & $\mathbf{0.0396 \pm 0.0514}$ \\
    \addlinespace
    \multirow{2}{*}{Mamba}
      & Baseline & $0.0950  \pm 0.0418$  & $0.0636 \pm 0.0774$ \\
      & Prop.    & $\mathbf{0.0358 \pm 0.0081}$ & $\mathbf{0.0495 \pm 0.0609}$ \\
    \bottomrule
  \end{tabular}
\end{table}
 

\section{CONCLUSIONS}
\label{sec:conclusion}
    We proposed Impedance Cloning, an imitation learning framework for
    contact-rich tasks that redefines the action space using stiffness $K^{\mathrm{eq}}$ and
    equilibrium point $\theta^{\mathrm{eq}}$ based on the EP hypothesis.
    We first confirmed that these parameters can be estimated from bilateral teleoperation
    demonstrations, and then validated the framework across three manipulation tasks: wiping,
    pick-and-place, and grasping.
    Experimental results demonstrated that the proposed method maintains stable contact forces
    under height perturbations where the baseline fails, adapts to unseen object geometries, and
    improves torque tracking accuracy when integrated with generative sequence modeling
    architectures (ILBiT and Mamba).

    These outcomes show that taking upper-level parameters grounded in biomechanical insight as
    the learning target is effective for robust contact-rich manipulation. We also showed that the
    framework can be formulated in both joint and Cartesian spaces, and that, in the Cartesian-space
    formulation, the proposed method is less sensitive than fixed-impedance replay to positive
    height offsets. While we do not claim to measure
    muscle-level stiffness, our results validate that extracting and imitating the operator's
    latent command parameters provides sufficient adaptability for real-world uncertainties.
    Future work includes extending this method to dynamic tasks with multi-directional
    contacts, and examining how to choose between the two coordinate spaces, including tasks that
    require exploiting the null space.

\end{document}